\RequirePackage[2020-02-02]{latexrelease}

\documentclass[a4paper,conference]{IEEEtran}
\usepackage[printwatermark]{xwatermark}
\newwatermark*[page=1,color=gray!60,fontfamily=cmr,fontseries=m,scale=0.4,xpos=0,ypos=140,width=2\paperwidth]{This paper has been accepted for publication at the \\ 26th International Conference on Pattern Recognition (ICPR), Montr\'eal Qu\'ebec, 2022}

\usepackage{multirow}
\usepackage{graphicx}
\usepackage[table, dvipsnames]{xcolor}
\definecolor{myblue}{rgb}{0.82, 0.95, 0.93}
\ifCLASSINFOpdf
\else
\fi
\usepackage{amsmath}
\usepackage{amssymb}
\begin{document}
%
\title{Kernel Modulation: A Parameter-Efficient Method for Training Convolutional Neural Networks}

\author{
Yuhuang Hu\qquad Shih-Chii Liu \\
Institute of Neuroinformatics, University of Z\"urich and ETH Z\"urich, Switzerland \\
\texttt{\{yuhuang.hu, shih\}.ini.uzh.ch}
}


%


\maketitle

\begin{abstract}
Deep Neural Networks, particularly Convolutional Neural Networks (ConvNets), have achieved incredible success in many vision tasks, but they usually require millions of parameters for good accuracy performance. With increasing applications that use ConvNets, updating hundreds of networks for multiple tasks on an embedded device can be costly in terms of memory, bandwidth, and energy. Approaches to reduce this cost include model compression and parameter-efficient models that adapt a subset of network layers for each new task.
This work proposes a novel parameter-efficient kernel modulation (KM) method that adapts all parameters of a base network instead of a subset of layers.
KM uses lightweight task-specialized kernel modulators that require only an additional 1.4\% of the base network parameters. With multiple tasks, only the task-specialized KM weights are communicated and stored on the end-user device. We applied this method in training ConvNets for Transfer Learning and Meta-Learning scenarios. Our results show that KM delivers up to 9\% higher accuracy than other parameter-efficient methods on the Transfer Learning benchmark.
\end{abstract}


%
\IEEEpeerreviewmaketitle

\section{Introduction}

Convolutional Neural Networks (ConvNets) have been successfully applied to many Computer Vision tasks such as object recognition and detection at the expense of over-parameterization~\cite{alex:net:Krizhevsky:2012, resnet:He:2016}.
As the need for fast adaptation and customization of ConvNets grows, the challenge for storing and distributing these large models arises.

One common strategy of addressing this challenge is by developing smaller ConvNets, and therefore reducing the memory requirement of the target platforms.
Methods to produce smaller networks include compression of large trained networks (comprehensive review in~\cite{network:compression:survey:Deng:2020}) and design of run-time efficient models such as ShuffleNetV2~\cite{shuffle:net:Ma:2018},  MobileNetV3~\cite{mobilenet:v3:Howard:2019}, and EfficientDet~\cite{efficient:det:Tan:2020}.
However, with the rapid increase of task-specialized networks, maintaining and updating a large number of these networks will still incur high memory storage cost and energy.

\begin{figure}[ht]
    \centering
    \includegraphics[width=\linewidth]{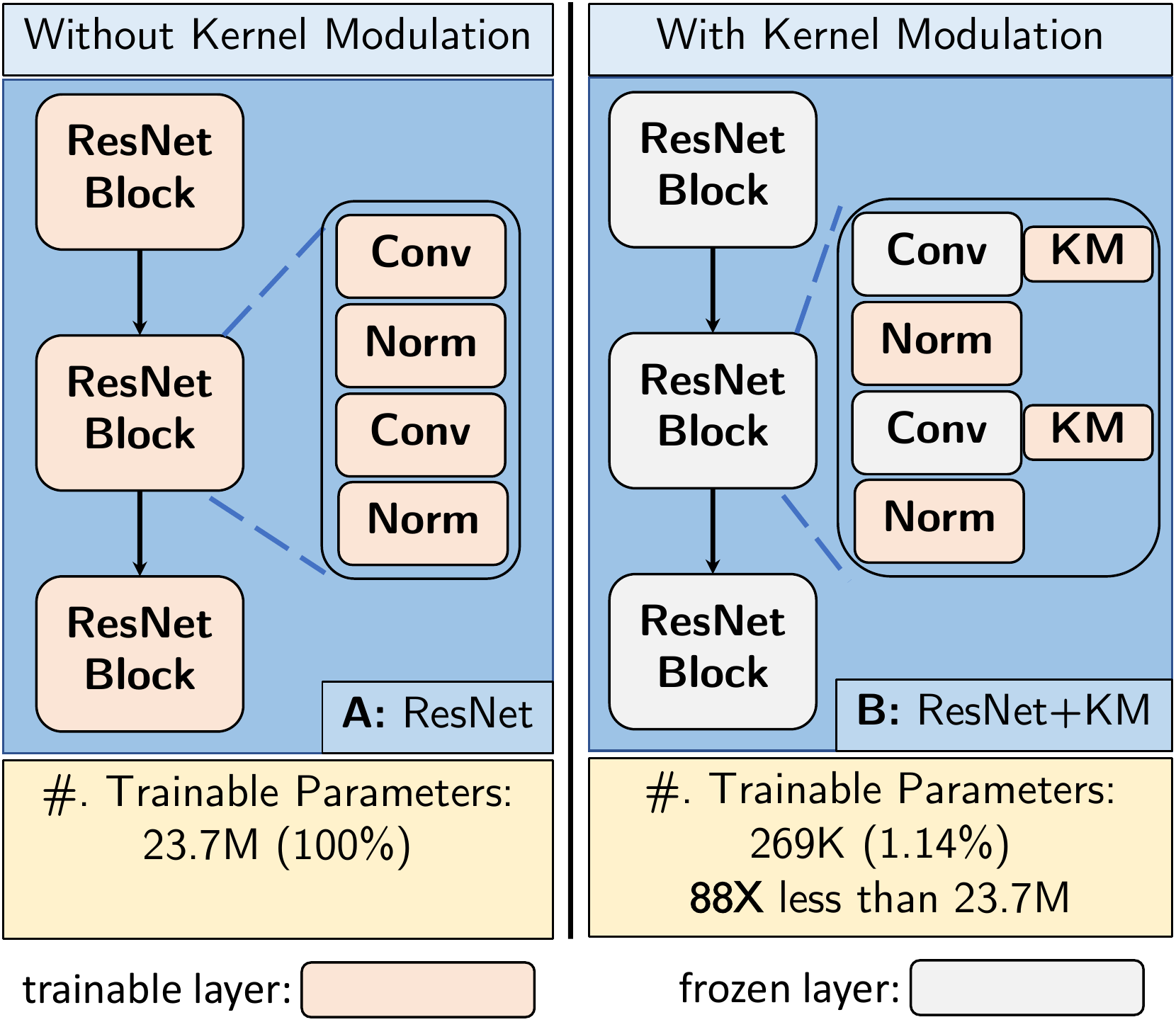}
    \caption{\textbf{A:} In standard training, all parameters of the ResNet-50 are trainable.
    \textbf{B:} With Kernel Modulation (KM), the convolution weights are frozen. Only the KM weights are trainable.}
    \label{fig:kernel:modulation:front:page}
\end{figure}

Various methods have been proposed for optimal training of networks for new tasks as explored in Transfer Learning and Meta-Learning scenarios~\cite{transfer:learning:Kornblith:2019, maml:Finn:2017}. Because training all parameters of a base network require large training resources and the storage of large parameter sets for multiple networks, parameter-efficient methods avoid fine-tuning of all parameters, instead, they adapt only certain layers of the network~\cite{train:batchnorm:for:transfer:Mudrakarta:2019, anil:Raghu:2020}. This translates to the memory required only for storing a subset of parameters, which is desirable for sharing these networks.

We propose a novel parameter-efficient method, 
the Kernel Modulation (\textbf{KM}) method which freezes all convolution weights in the base network during task adaptation. Instead, task-specialized lightweight \emph{kernel modulators} are trained to produce the modulated convolution weights as outlined in Fig.~\ref{fig:kernel:modulation:front:page}.
These small kernel modulator networks are much smaller than the base network.
The contributions of this work are as follows:
\begin{itemize}
    \item We propose the KM method for task adaptation that requires, on average, only an additional 1.4\% of the base network parameters for each task that translates to a decrease of \textbf{71X} in memory footprint.
    \item We introduce two forms of modulation: explicit modulation (Sec.~\ref{subsec:method:km}) and implicit modulation (Sec.~\ref{subsec:method:norm}). \emph{Explicit} modulation directly adapts the weights via the kernel modulators implemented as Multi-layer Perceptrons (MLPs). \emph{Implicit} modulation is carried out by feature normalization layers~\cite{batch:norm:Ioffe:2015, group:norm:Wu:2018} that indirectly modulates the weights through affine transformation on the network activations, \emph{i.e.}, feature maps.
\end{itemize}

Our experiments show that the KM method surpasses other parameter-efficient training methods significantly when training a ConvNet on a classification task (Sec.~\ref{subsec:km:random:network:modulation}).
When tested in both Transfer Learning and Meta-Learning scenarios, KM achieves better or comparable accuracy (up to 9\%) across multiple benchmark datasets when compared with other methods (Sec.~\ref{subsec:km:transfer}, \ref{subsec:km:meta}). 
Empirically, we also found that the KM models generalize well, \emph{i.e.}, they do not overfit on new tasks because the number of trainable parameters is much smaller than the number of available training samples.

\section{Related Work}\label{sec:related:works}

Using a small number of trainable parameters when optimizing a large network is attractive for deployment on multiple platforms~\cite{train:batchnorm:for:transfer:Mudrakarta:2019} and helps a better understanding of the intrinsic dimension of a large neural network~\cite{low:dimension:training:Li:2018}. The authors in
\cite{all:layers:created:equal:Zhang:2019} showed that many convolution layers in ResNets are ``ambient'' layers where the weights stay close to their initialized values throughout the training, in contrast to ``critical'' layers. These findings suggest that many tasks could be solved by training only a subset of the parameters or by small modulations of the initial parameters.

Parameter-efficient training methods in the literature can be roughly divided into two related groups.
One, by \emph{training a low-dimensional parameter vector}
as pioneered by~\cite{low:dimension:training:Li:2018} and \cite{low:dimension:training:Gressmann:2020}. 
Here, the ConvNet weights are updated via a set of random bases that project the trained low-dimensional parameters to the high-dimensional weight space. The second group consists of methods for \emph{training a subset of weights in the network}.
A common practice for new task adaptation is to fine-tune only the last layer of the network during task adaptation~\cite{transfer:learning:Donahue:2014, transfer:learning:Cui:2018}
or a subset of layers (normalization and convolution layers)~\cite{train:batchnorm:for:transfer:Mudrakarta:2019}.
More recently, \cite{only:train:batchnorm:Frankle:2021} showed that training only Batch Normalization (BN) layers~\cite{batch:norm:Ioffe:2015} even with random weights for the remaining layers, led to a surprisingly high classification accuracy (69.5\% for CIFAR-10 using ResNet-101).
Our KM method differs from previous methods by modulating \emph{all} parameters during training while using a small number of trainable parameters.

The explicit modulation in our method bears some similarity to the elastic kernel transformation method used in Neural Architecture Search (NAS)~\cite{once:for:all:Cai:2020}. A kernel transformation matrix to accommodate different kernel sizes, is trained during the search. After training, the transformation matrix is used to implement networks that are suitable for a network specification, \emph{e.g.}, inference speed and model size.

\section{Methods}

Sec.~\ref{subsec:method:km} presents the explicit kernel modulation of the convolution weights. Sec.~\ref{subsec:method:norm} discusses how normalization layers, such as Batch Normalization and Group Normalization layers frequently used in ConvNets, can be interpreted as a form of implicit kernel modulation. Sec.~\ref{subsec:method:dataset} summarizes the datasets used in this work.

\subsection{Explicit Kernel Modulation} \label{subsec:method:km}

A ConvNet $f$ consists of a number of convolution layers that are parameterized by $N$ weights, $\mathbf{W}\in\mathbb{R}^{N}$. The network takes an input $\mathbf{x}$ and outputs $\mathbf{y}$:
\begin{equation}
    \mathbf{y} = f(\mathbf{x}; \mathbf{W})
\end{equation}

Instead of training all weights through stochastic gradient descent, we introduce an additional kernel modulator $g$ that regulates $\mathbf{W}$.
The modulator $g$ has $M$ parameters $\mathbf{U}\in\mathbb{R}^{M}$ where $M\ll N$.
\begin{equation}
    \mathbf{y}=f(\mathbf{x}; \tilde{\mathbf{W}}); \quad \tilde{\mathbf{W}}=g(\mathbf{W}; \mathbf{U}) \label{eqn:km:formulation}
\end{equation}
In Eq.~\ref{eqn:km:formulation}, the kernel modulator $g$ takes $\mathbf{W}$ as input and outputs modulated convolution weights $\tilde{\mathbf{W}}$. By only training $\mathbf{U}$, The explicit kernel modulation avoids training of the convolution weights of the ConvNet $f$.

\begin{figure}[ht]
    \centering
    \includegraphics[width=\linewidth]{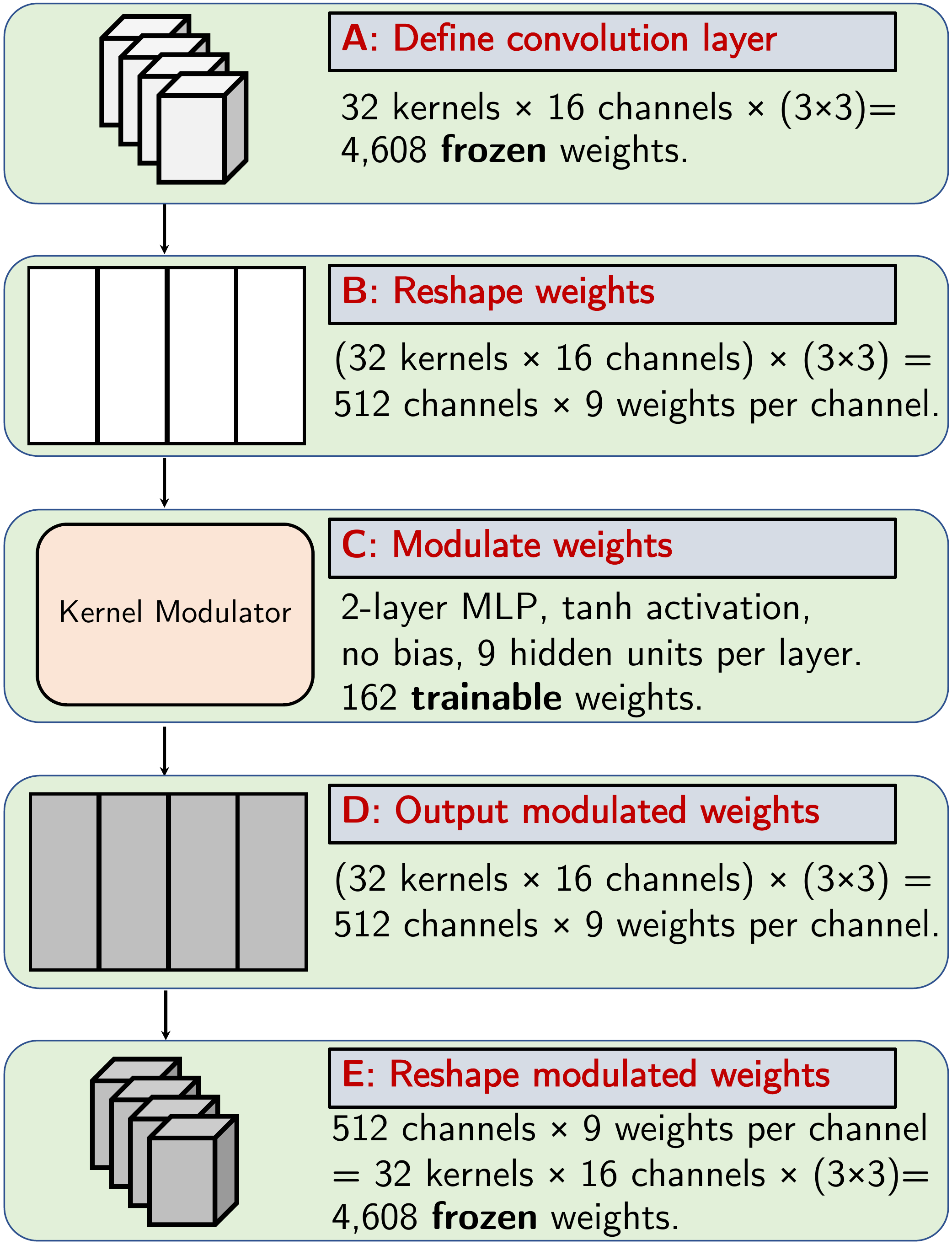}
    \caption{\textbf{Explicit Kernel Modulation (KM) steps.} Sub-figures \textbf{A}-\textbf{E} illustrate the steps to modulate a ResNet-32 convolution layer that has 32 kernels of size $3\times 3$, and each kernel has 16 channels.}
    \label{fig:kernel:modulation:steps}
\end{figure}

We perform explicit modulation at each convolution layer $f^{(i)}$ via its own kernel modulator $g^{(i)}$ where $(i)$ is the index of the layer. The explicit KM consists of five steps (A-E) that are illustrated in Fig.~\ref{fig:kernel:modulation:steps}:
\begin{enumerate}
    \item[\textbf{A.}] Let $f^{(i)}$'s convolution weights, $\mathbf{W}^{(i)}$, be frozen. There are $k_{n}$ kernels. Each kernel has $k_{c}$ channels and a spatial dimension $k_{h}\times k_{w}$, \emph{i.e.}, the shape is $(k_{n}, k_{c}, k_{h}, k_{w})$.
    \item[\textbf{B.}] We reshape $\mathbf{W}^{(i)}$ from a 4D tensor to a 2D tensor where its dimension is $(k_{n}\times k_{c}, k_{h}\times k_{w})$. In the example shown in Fig.~\ref{fig:kernel:modulation:steps}, the dimension of the convolution weights is reshaped from (32, 16, 3, 3) to (512, 9).
    \item[\textbf{C.}] The corresponding kernel modulator $g^{(i)}$ is implemented as a 2-layer MLP network that has both $k_{h}\times k_{w}$ inputs and outputs. $g^{(i)}$'s parameters, $\mathbf{U}^{(i)}$, are trainable.
    \item[\textbf{D.}] The network $g^{(i)}$ takes $\textbf{W}^{(i)}$ as input and returns modulated weights $\tilde{\mathbf{W}}^{(i)}$.
    \item[\textbf{E.}] The modulated weights $\tilde{\mathbf{W}}^{(i)}$ are reshaped back to its original shape $(k_{n}, k_{c}, k_{h}, k_{w})$.
\end{enumerate}

The modulated weights $\tilde{\mathbf{W}}^{(i)}$ are used in the convolution layer $f^{(i)}$ for transforming the input $\mathbf{x}^{(i)}$ into its output $\mathbf{y}^{(i)}$:
\begin{equation}
    \mathbf{y}^{(i)}=f^{(i)}(\mathbf{x}^{(i)}; \tilde{\mathbf{W}}^{(i)}); \quad \tilde{\mathbf{W}}^{(i)}=g^{(i)}(\mathbf{W}^{(i)}; \mathbf{U}^{(i)})
\end{equation}
Note that only the kernel modulator's weights $\mathbf{U}^{(i)}$ are trainable. In the example shown in Fig.~\ref{fig:kernel:modulation:steps}, compared to training 4.6K convolution weights, KM only trains 162 parameters ($\sim$28X reduction).

\paragraph{Kernel modulator initialization} The kernel modulator $g^{(i)}$ is a 2-layer MLP network that uses the $\tanh$ activation function. Each layer's weight is initialized as
\begin{equation}
    \mathbf{U}^{(i, j)}=\mathbf{I}+\mathcal{N}(0, 0.001); \quad j=1, 2 \label{eqn:km:init}
\end{equation}
where $\mathbf{U}^{(i,j)}$ denotes the weights of the $j$-th hidden layer of $g^{(i)}$, $\mathbf{I}$ is the identity matrix; $\mathcal{N}(\mu, \sigma)$ is a normal distribution with $\mu=0$ and $\sigma=0.001$.

\begin{figure}[ht]
    \centering
    \includegraphics[width=\linewidth]{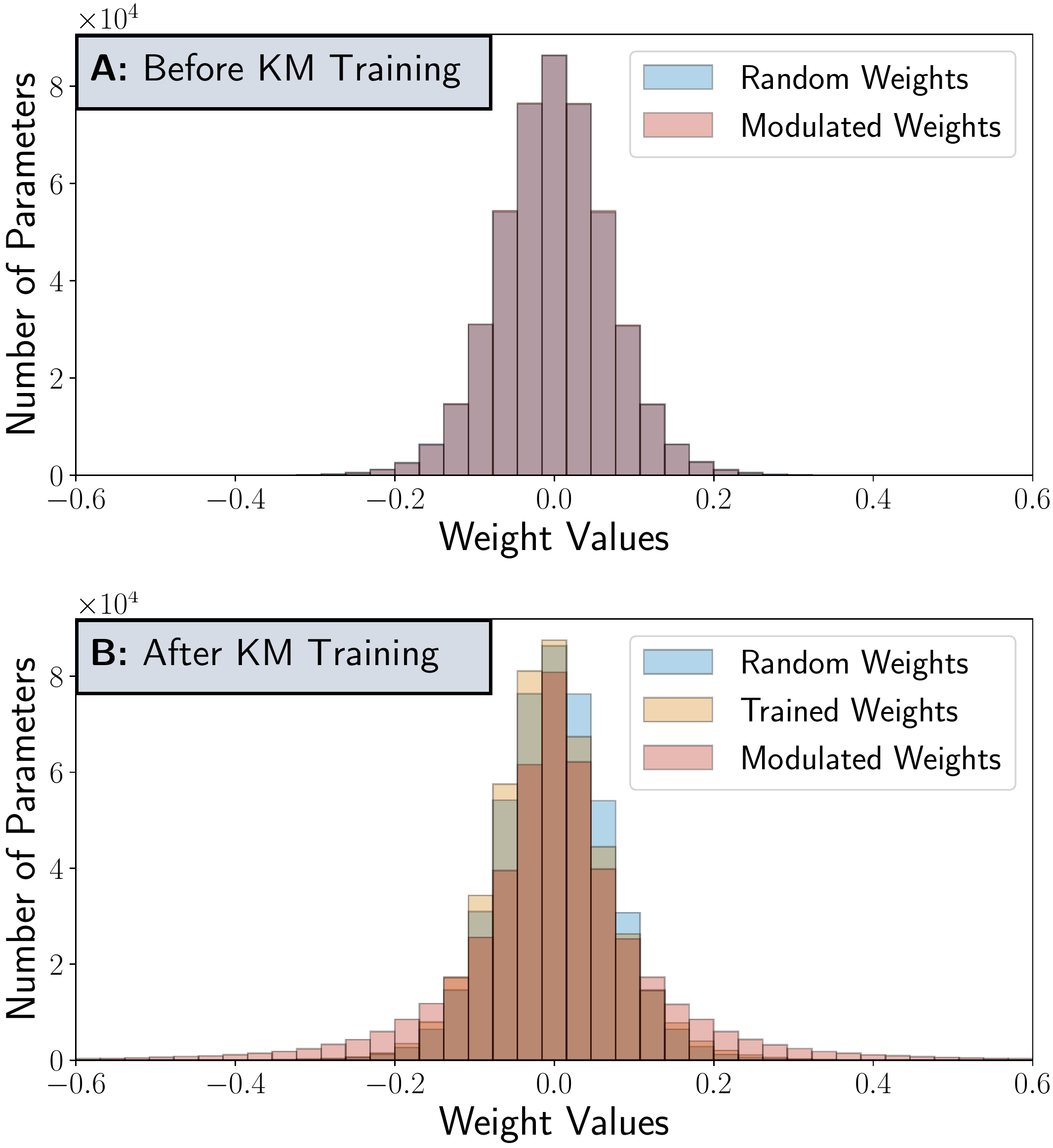}
    \caption{\textbf{Comparing weights of a ResNet-32 network trained on CIFAR-10 dataset.} \textbf{A}: Before KM training, the modulated weights preserve the random convolution weight values; \textbf{B}: After KM training, the modulated weights have a wider distribution than the original convolution weights. The trained weights of a network in which all parameters were trained are also shown here for comparison.}
    \label{fig:compare:conv:km:weights}
\end{figure}

We use the $\tanh$ activation function and the initialization method in Eq.~\ref{eqn:km:init} to preserve the convolution weight values before modulation (Fig.~\ref{fig:compare:conv:km:weights}A).
After training, the kernel modulator $g$ is used to regulate the original convolution weights (Fig.~\ref{fig:compare:conv:km:weights}B).
We study in Sec.~\ref{subsec:km:design:kernel:modulator}, the impact of choosing different activation functions, initialization methods, and the number of layers.

\subsection{Normalization as implicit kernel modulation}\label{subsec:method:norm}

Feature normalization layers such as Batch Normalization (BN) and Group Normalization (GN) are essential to ConvNets. We show here that the normalization layers, BN and GN, \emph{implicitly} modulate the convolution weights.

Let the input feature maps, $\mathbf{x}^{(i)}$, and output feature maps, $\mathbf{y}^{(i)}$, be of dimension $(B, C, H, W)$ where $B$ is the batch size, $C$ is the number of channels, $H$ and $W$ are spatial dimensions. We consider ResNet architectures where the normalization is followed by a convolution layer, \emph{i.e.}, $\mathbf{y}^{(i)}=\text{Conv}(\text{Norm}(\mathbf{x}^{(i)}))$. The analysis of the reversed case where the convolution is followed by a normalization layer is presented in the supplementary material. This reversed case is commonly known as ``Normalization Folding'' when implemented during network inference~\cite{normalization:folding:Rueckauer:2017}.

Output feature maps $\mathbf{y}^{(i)}$ can be computed using the associated convolution kernels $\mathbf{W}^{(i)}$ and the normalized input feature maps $\hat{\mathbf{x}}^{(i)}$ (the bias is omitted for brevity):
\begin{equation}
    \mathbf{y}^{(i)} = \mathbf{W}^{(i)}\otimes\hat{\mathbf{x}}^{(i)} \label{eqn:convolution}
\end{equation}
where `$\otimes$' denotes the convolution, and:
\begin{equation}
    \hat{\mathbf{x}}^{(i)}=\frac{\gamma}{\sigma}(\mathbf{x}^{(i)}-\mu)+\beta \label{eqn:norm:input}
\end{equation}
The running mean and standard deviation $\mu$ and $\sigma$ are calculated from $\mathbf{x}^{(i)}$. For instance, BN calculates channel-wise $\mu$ and $\sigma$. The trainable parameters $\gamma$ and $\beta$ have $C$ values, one for each feature channel. Substituting Eq.~\ref{eqn:norm:input} into Eq.~\ref{eqn:convolution}:
\begin{align}
    \mathbf{y}^{(i)}=&\mathbf{W}^{(i)}\otimes\left(\frac{\gamma}{\sigma}(\mathbf{x}^{(i)}-\mu)+\beta\right) \\
    =&\frac{\gamma}{\sigma}\mathbf{W}^{(i)}\otimes\mathbf{x}^{(i)}+\mathbf{W}^{(i)}\otimes\left(\beta-\frac{\gamma}{\sigma}\mu\right) \label{eqn:norm:modulation}
\end{align}
From Eq.~\ref{eqn:norm:modulation}, we can interpret the normalization layer that modulates the convolution kernels $\mathbf{W}^{(i)}$ via a factor $\gamma/\sigma$ and an offset $\mathbf{W}^{(i)}\otimes(\beta-\gamma\mu/\sigma)$ while Ref.~\cite{only:train:batchnorm:Frankle:2021} viewed BN as applying an affine transformation to the network activation.

\subsection{Datasets} \label{subsec:method:dataset}

We used the CIFAR-10 dataset~\cite{sl:dataset:cifar10:Krizhevsky:2009} for the KM experiments in Sec.~\ref{subsec:km:random:network:modulation} and
the following four datasets: Aircraft~\cite{tl:dataset:aircraft:Maji:2013}, Cars~\cite{tl:dataset:cars:Krause:2013}, Flowers~\cite{tl:dataset:flowers:Nilsback:2008}, and Foods~\cite{tl:dataset:foods:Bossard:2014} for the Transfer Learning experiments in  Sec.~\ref{subsec:km:transfer}.
Two datasets, Omniglot~\cite{meta:dataset:omniglot:Lake:2011} and mini-ImageNet~\cite{meta:dataset:miniimagenet:Ravi:2017}, are used for the Meta-Learning experiments in Sec.~\ref{subsec:km:meta}. Detailed information about the datasets are provided in the supplementary material.

\begin{table*}[ht]
    \caption{Comparison of KM with other parameter-efficient methods when training a ResNet-32 from scratch on CIFAR-10. KM results are highlighted in blue. `Y' means that the respective component is included during training.}
    \label{tab:resnet32:cifar10}
    \centering
    \begin{tabular}{c|l|cccc|c|c|l|l}
        \hline
        \multirow{2}{*}{\textbf{Model}} &\multicolumn{1}{c|}{\multirow{2}{*}{\textbf{Network}}} &  \multicolumn{4}{c|}{\textbf{Training Combinations}} & \multicolumn{1}{c|}{\multirow{1}{*}{\textbf{\#~Trainable}}} & \multicolumn{1}{c|}{\multirow{1}{*}{\textbf{\#~Total}}} &
        \multicolumn{1}{c|}{\textbf{Accuracy}} & \multicolumn{1}{c}{\textbf{Recovered}}\\
        \cline{3-6}
        & & Convolution & Implicit & Explicit & Classifier & \multicolumn{1}{c|}{\multirow{1}{*}{\textbf{Parameters}}} & \multicolumn{1}{c|}{\multirow{1}{*}{\textbf{Parameters}}} & \multicolumn{1}{c|}{\textbf{(\%)}} & \multicolumn{1}{c}{\textbf{Accuracy Ratio}} \\
        \hline
        \texttt{BL} & ResNet-32 & & Y & & Y & 2.9\,K & 470\,K & 59.58\textsuperscript{$\pm$1.44} & \textcolor{blue}{\rule{1.28cm}{1mm}} 0.64\\
        \hline
        \rowcolor{myblue}
        \texttt{KM} & ResNet-32 & & Y & Y & Y & 7.9\,K & 470\,K & \textbf{77.60\textsuperscript{$\pm$0.44}} & \textcolor{red}{\rule{1.68cm}{1mm}} 0.84\\
        \hline
        \texttt{SN} & Small ResNet~\cite{only:train:batchnorm:Frankle:2021} & Y & Y & & Y & 8.0\,K & 8.0\,K & 71.00 & \textcolor{blue}{\rule{1.54cm}{1mm}} 0.77\\
        \hline
        \hline
        \texttt{LD} & \multicolumn{5}{l|}{RBD~\cite{low:dimension:training:Gressmann:2020}} & 8.0\,K & 78.3\,K & 70.26\textsuperscript{$\pm$0.02} & \textcolor{blue}{\rule{1.52cm}{1mm}} 0.76\\
        \texttt{TBN} &\multicolumn{5}{l|}{ResNet-110~\cite{only:train:batchnorm:Frankle:2021}} & 8.3\,K & 1.7\,M & 69.50 & \textcolor{blue}{\rule{1.5cm}{1mm}} 0.75\\
        \hline
    \end{tabular}
\end{table*}

\section{Results}\label{sec:experiments}

We demonstrate Kernel Modulation in three sets of experiments described in the next 3 subsections: 1) Training a ConvNet  from scratch; 2) Transfer Learning; and 3) Meta-Learning. Sec.~\ref{subsec:km:design:kernel:modulator} studies how different constructions of kernel modulators impact the model accuracy. All experiments are repeated for five runs with different random seeds.

\subsection{Training a ConvNet from scratch} \label{subsec:km:random:network:modulation}

Starting with random weights for the network, 
we compare the accuracy of using KM on a ResNet-32 against other parameter-efficient methods~\cite{low:dimension:training:Gressmann:2020, only:train:batchnorm:Frankle:2021} on an object recognition task; and to the accuracy of training the entire ResNet-32~\cite{resnet:He:2016}; . We use the original ResNet-32 training schedule~\cite{resnet:He:2016} where each model is trained for 200 epochs with batch size 128. All networks are trained on CIFAR-10.
By training all parameters, the accuracy of a ResNet-32 reached 92.8\%.
For comparing across different parameter-efficient methods, we define a metric \emph{Recovered Accuracy Ratio} $\in[0,1]$ as the ratio between the accuracy achieved by a parameter-efficient method and the accuracy of the fully trained ResNet-32. A larger ratio indicates a better method.

We compare the \texttt{KM} model with a baseline model \texttt{BL} model where only the implicit modulation (BN layers) and the linear classifier are trained.
The results in Table~\ref{tab:resnet32:cifar10} show that combined with explicit modulation, the \texttt{KM} model can significantly increase the recovered accuracy ratio by 20\% at a cost of 5\,K additional parameters.
We also compare the \texttt{KM} model with a small ResNet \texttt{SN}~\cite{only:train:batchnorm:Frankle:2021} which consists of a similar number of trainable parameters. Despite training all parameters of the \texttt{SN} model, the recovered accuracy ratio is still lower than the \texttt{KM}'s accuracy by 7\%.


We further compare the KM model with two other parameter-efficient training methods. The Random Bases Descent (RBD) model, \texttt{LD}, trains a low-dimensional trainable parameter vector and updates the network via a set of random bases that are sampled at every training iteration. The \texttt{TBN} model is similar to \texttt{BL} where only normalization layers and the classifier of a ResNet-110 are trained.
The \texttt{KM} model adapts all network parameters and the recovered accuracy ratio surpasses that of the RBD model by 8\% and the \texttt{TBN} model by 9\%.

\subsection{Transfer Learning} \label{subsec:km:transfer}

Kernel modulators are trained while the original convolution weights are frozen. After training, these task-specialized, lightweight kernel modulators are distributed instead of sharing the entire base network for Transfer Learning. 
We evaluated KM in Transfer Learning experiments on four benchmark datasets: Aircraft, Cars, Flowers and Food (Sec.~\ref{subsec:method:dataset}).
The base network is a ResNet-50~\cite{resnet:He:2016} backbone that is pretrained on the ImageNet dataset~\cite{imagenet:Deng:2009} and has 23.5\,M parameters. For each task, we re-initialize the linear classifier according to the number of classes of the target dataset. All networks are trained for 50 epochs with a batch size of 8.
Different learning rates are chosen for different datasets (see Table~\ref{tab:transfer:learning:result}). To accommodate a small batch size during KM, we replace BN with GN.

\begin{figure}[ht]
    \centering
    \includegraphics[width=\linewidth]{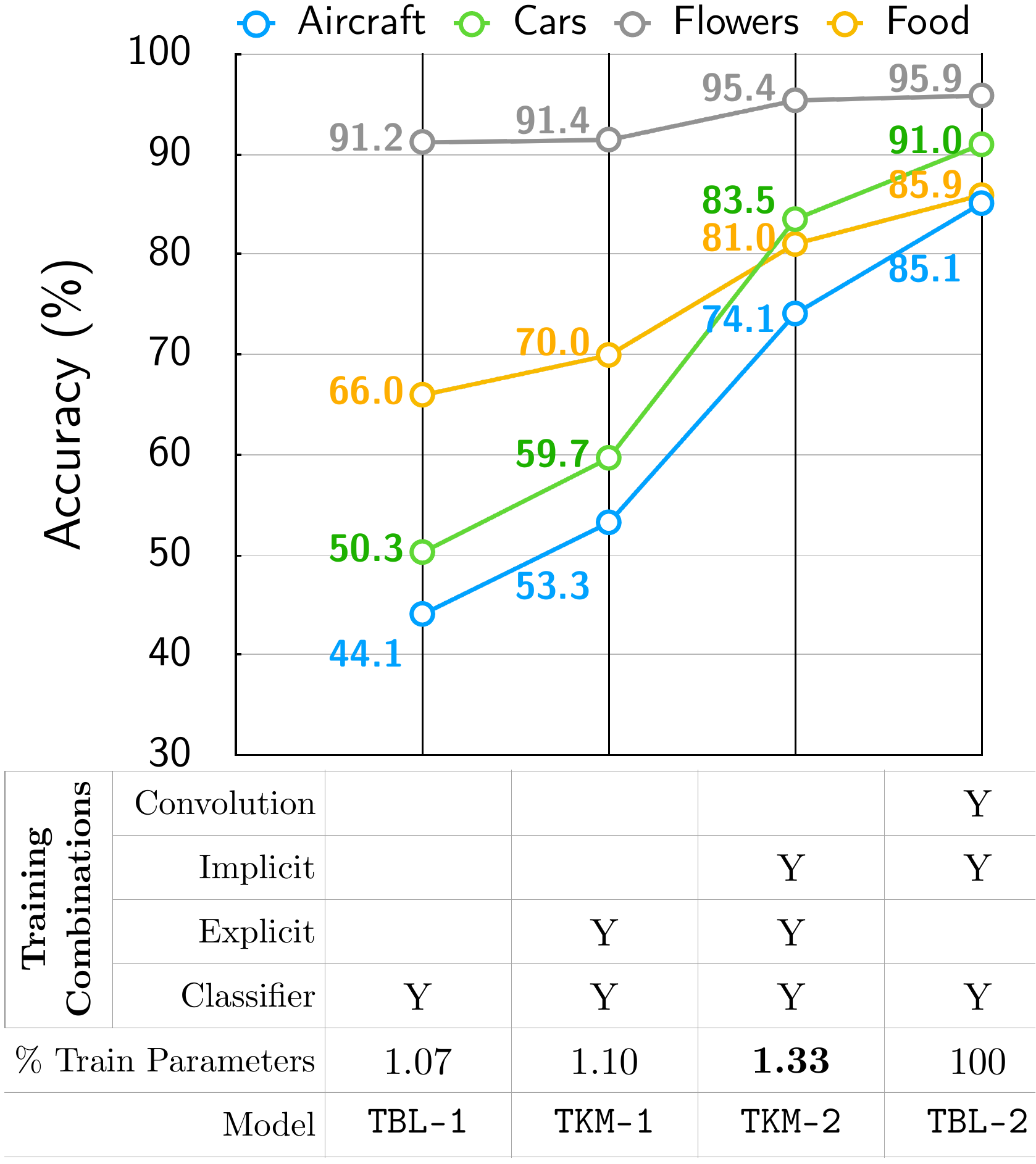}
    \caption{\textbf{Transfer Learning Results.} Each column of the table is a combination of modules where the weights are adapted during training. ``Y'' means that the corresponding component is included.}
    \label{fig:trans:learning:results}
\end{figure}

\begin{table*}[ht]
    \centering
    \caption{Comparison of KM with other Transfer Learning methods on four benchmark datasets. \texttt{NC-1,2}, \texttt{TTL} models are parameter-efficient methods for Transfer Learning. \texttt{TTL}, and \texttt{TBL-2} models fine-tune all parameters in the network. KM results are highlighted in blue.}
    \label{tab:transfer:learning:result}
    \begin{tabular}{c|p{1.5cm}|l|llll}
        \hline
        \multicolumn{1}{c|}{\multirow{2}{*}{\textbf{Model}}}& \multicolumn{1}{c|}{\multirow{2}{*}{\textbf{Method}}} & \textbf{Dataset} & Aircraft & Cars & Flowers & Food \\
        \cline{3-7}
        & & \textbf{Learning Rate} & 16e-4 & 8e-4 & 2e-4 & 2e-4 \\
        \hline
        \texttt{NC-1} & \multicolumn{2}{l|}{Implicit+Classifier~\cite{train:batchnorm:for:transfer:Mudrakarta:2019}} & 70.70 & 81.00 & 90.40 & --- \\
        \texttt{NC-2} & \multicolumn{2}{l|}{Implicit+Classifier~\cite{tinytl:Cai:2020}} & 68.10 & 77.90 & 94.30 & 77.00 \\
        \texttt{TTL} & \multicolumn{2}{l|}{TinyTL-L+B~\cite{tinytl:Cai:2020}} & 75.40 & 85.00 & 95.50 & 79.70 \\
        \hline
        \rowcolor{myblue}
        \texttt{TKM-1} & \multicolumn{2}{l|}{Explicit+Classifier (Ours)} & 53.26\textsuperscript{$\pm$0.51} & 59.69\textsuperscript{$\pm$0.23} & 91.45\textsuperscript{$\pm$0.05} & 69.96\textsuperscript{$\pm$0.22} \\
        \rowcolor{myblue}
        \texttt{TKM-2} & \multicolumn{2}{l|}{Implicit+Explicit+Classifier (Ours)} & 74.09\textsuperscript{$\pm$0.58} & 83.51\textsuperscript{$\pm$0.27} & 95.37\textsuperscript{$\pm$0.17} & 81.03\textsuperscript{$\pm$0.16} \\
        \hline
        \texttt{FTL} & \multicolumn{2}{l|}{All Parameters~\cite{transfer:learning:Kornblith:2019}} & 86.60 & 91.70 & 97.51 & 87.80 \\
        \texttt{TBL-2} & \multicolumn{2}{l|}{All Parameters (Ours)} & 85.12\textsuperscript{$\pm$0.67} & 90.98\textsuperscript{$\pm$0.17} & 95.85\textsuperscript{$\pm$0.23} & 85.89\textsuperscript{$\pm$0.18}\\
        \hline
    \end{tabular}
\end{table*}

In Fig.~\ref{fig:trans:learning:results}, we established two commonly used Transfer Learning baselines. The first is obtained by using a pretrained backbone as the feature extractor and training only the classifier  (\texttt{TBL-1})~\cite{transfer:learning:Donahue:2014, transfer:learning:Cui:2018}. The second is by fine-tuning all parameters (\texttt{TBL-2})~\cite{transfer:learning:Kornblith:2019, train:batchnorm:for:transfer:Mudrakarta:2019}. KM training is conducted in two combinations: 1) explicit modulation only (\texttt{TKM-1}); 2) explicit and implicit modulations (\texttt{TKM-2}).
When the new dataset is most dissimilar from the backbone's pretrained dataset, such as Aircraft and Cars datasets, training only the classifier as in the \texttt{TBL-1} model does not lead to good transfer accuracy. It is also necessary to adapt the backbone's parameters. After applying explicit modulation, \texttt{TKM-1} model improves the average accuracy by 11\% over that of the \texttt{TBL-1} model. With implicit modulation, the accuracy of \texttt{TKM-2} model is 40\% higher than that of the \texttt{TBL-1} model even with 75X fewer trainable parameters than the \texttt{TBL-2} model.

Table~\ref{tab:transfer:learning:result} compares the accuracy of our KM model with two parameter-efficient Transfer Learning models: 1) in the \texttt{NC-1, NC-2} models where normalization layers (implicit modulation) and the final classification layer are fine-tuned; 2) in the \texttt{TTL} model where additional layers are added alongside the original network for tuning network activation. The \texttt{TKM-2}'s accuracy surpassed \texttt{NC-1, NC-2}'s accuracy on all datasets.
Our model accuracy is also comparable to the state-of-art accuracy of the \texttt{TTL} model without the addition of side branches. Results show clearly that both explicit and implicit modulations are needed for attaining good transfer accuracy (\texttt{TKM-1} \emph{vs.} \texttt{TKM-2}). The highest accuracy for Transfer Learning tasks is achieved by fine-tuning all parameters in the network (\texttt{FTL}, \texttt{TBL-2}).

\subsection{Meta-Learning} \label{subsec:km:meta}

Both Transfer Learning and Meta-Learning studies aim to adapt a pretrained base network to new tasks. The difference is that
Meta-Learning focuses on task adaptation using a small number of training samples. Here, we study the use of KM in the Model-Agnostic Meta-Learning (MAML)~\cite{maml:Finn:2017} framework which has two training loops. The outer loop pretrains the backbone network for a ``meta-initialization'' and the inner loop adapts the network for a new task by fine-tuning both the backbone and a task-specific classifier. The authors in~\cite{anil:Raghu:2020} proposed an alternative algorithm called Almost No Inner Loop (ANIL), in which only the classifier is trained during task adaptation. With the KM method, we combine the training of the classifier together with our parameter-efficient task-specialized update for the backbone; and compare the accuracy performance of MAML and ANIL with the tasks described next.

We set up four few-shot recognition tasks using the Omniglot and mini-Imagenet datasets. A few-shot task is coded as $n$-way-$k$-shot in which $n$ is number of classes in this task and $k$ is the number of training samples per class. The network setup, training, and validation procedures follow the same as in~\cite{maml:Finn:2017, anil:Raghu:2020}. For brevity, the exact setup description is in the supplementary material.

The results in Table~\ref{tab:meta:learning} show that the KM model achieves higher accuracy than the ANIL model on all tasks. KM also gives higher accuracy than MAML in the Omniglot 20-way-1-shot and mini-ImageNet 5-way-5-shot cases even when using $<$ 51X trainable parameters. In contrast to ANIL, we found that modulating the weights of the backbone to adapt to new tasks is beneficial.

\begin{table}[ht]
    \centering
    \caption{Meta-Learning comparison among few-shot tasks. This table reports both the accuracy from the original MAML and ANIL implementations. The KM results are higlighted in blue.}
    \label{tab:meta:learning}
    \begin{tabular}{l|cc}
        \hline
        & \multicolumn{2}{c}{\textbf{Omniglot Accuracy (\%)}} \\
        \hline
        \multicolumn{1}{c|}{\textbf{Method}} & 20-way-1-shot & 20-way-5-shot \\
        \hline
        MAML~\cite{maml:Finn:2017} & 95.80\textsuperscript{$\pm$0.30} & 98.90\textsuperscript{$\pm$0.20} \\
        MAML~\cite{anil:Raghu:2020} & 93.70\textsuperscript{$\pm$0.70} & 96.40\textsuperscript{$\pm$0.10} \\
        ANIL~\cite{anil:Raghu:2020} &  96.20\textsuperscript{$\pm$0.50} & 98.00\textsuperscript{$\pm$0.30} \\
        \hline
        ANIL (Ours) & 94.94\textsuperscript{$\pm$0.55} & 97.55\textsuperscript{$\pm$0.29}\\
        \rowcolor{myblue}
        KM (Ours) & 97.01\textsuperscript{$\pm$0.01} & 98.10\textsuperscript{$\pm$0.02} \\
        \hline
        \hline
        & \multicolumn{2}{c}{\textbf{mini-ImageNet Accuracy (\%)}} \\
        \hline
        \multicolumn{1}{c|}{\textbf{Method}} & 5-way-1-shot & 5-way-5-shot \\
        \hline
        MAML~\cite{maml:Finn:2017} & 48.70\textsuperscript{$\pm$1.84} & 63.11\textsuperscript{$\pm$0.92} \\
        MAML~\cite{anil:Raghu:2020} & 46.90\textsuperscript{$\pm$0.20} &  63.10\textsuperscript{$\pm$0.40} \\
        ANIL~\cite{anil:Raghu:2020} & 46.70\textsuperscript{$\pm$0.40} & 61.50\textsuperscript{$\pm$0.50} \\
        \hline
        ANIL (Ours) & 47.17\textsuperscript{$\pm$0.67} & 62.54\textsuperscript{$\pm$0.20} \\
        \rowcolor{myblue}
        KM (Ours) & 47.18\textsuperscript{$\pm$0.17} & 63.31\textsuperscript{$\pm$0.17} \\
        \hline
    \end{tabular}
\end{table}

\begin{figure*}[ht]
    \centering
    \includegraphics[width=\linewidth]{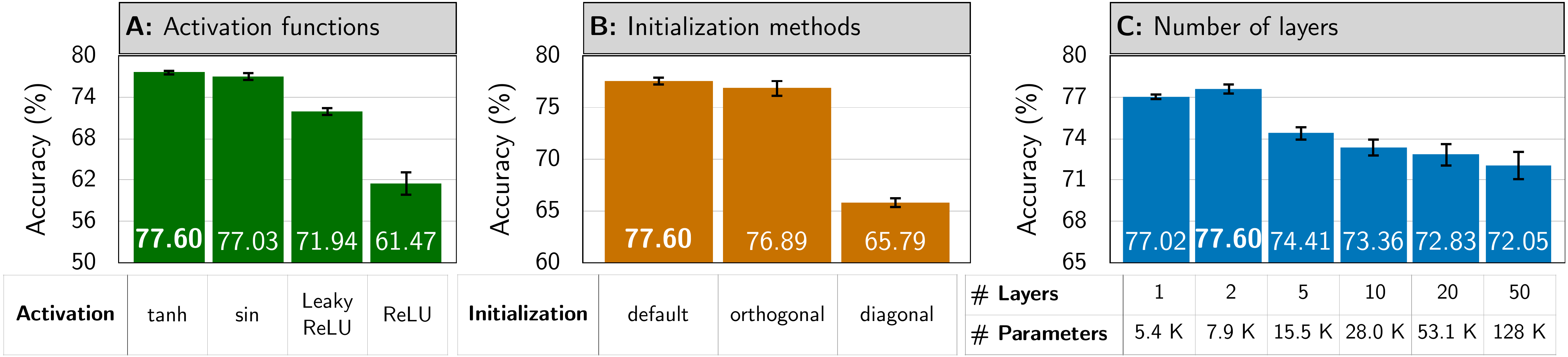}
    \caption{\textbf{Experiments with ResNet-32 and CIFAR-10 dataset.} \textbf{A}: Accuracy from using different activation functions on  2-layer MLP and the default initialization used ; \textbf{B}: Accuracy from using different initialization methods on 2-layer MLP and $\tanh$ activation; \textbf{C}: Accuracy from varying number of layers. $\tanh$ activation and the default initialization are used.}
    \label{fig:select:configs}
\end{figure*}

\subsection{Analysis of kernel modulator}
\label{subsec:km:design:kernel:modulator}

The experiments in the previous sections used a kernel modulator which consists of a 2-layer MLP network with a $\tanh$ activation function and initialization following Eq.~\ref{eqn:km:init} as defined in Sec.~\ref{subsec:method:km}.
In this section, we investigate the impact of three factors when constructing the kernel modulator, specifically, the type of activation function, the weight initialization method, and the number of layers. To avoid biases, we choose to train a ResNet-32 from scratch on a CIFAR-10 classification task following the same training schedule presented in Sec.~\ref{subsec:km:random:network:modulation}.

\paragraph{Activation function} The default activation function $\tanh$ is a symmetric function that allows both positive and negative outputs. We tested two symmetric activation functions: $\tanh$ and $\sin$ functions; and two asymmetric functions: Leaky ReLU and ReLU functions. The Leaky ReLU function has a negative slope of 0.1. As expected, the different symmetric activation functions led to similar accuracies  (Fig.~\ref{fig:select:configs}A).
Because the negative output values of the modulated weights are less weighted in asymmetric functions, their use leads to a significant accuracy drop and in addition, a large increase in the standard deviation of the accuracy values over multiple runs, thereby, confirming that the use of symmetric activation functions is desirable in KM.

\paragraph{Initialization methods} The default initialization in Eq.~\ref{eqn:km:init} preserves the input weight values.
We compare this method with two other initialization methods that could similarly keep the weight values.
The first candidate is the orthogonal method which initializes the kernel modulator's weights as orthogonal matrices~\cite{ortho:init:Saxe:2014}. The second candidate is diagonal initialization in which only the diagonal values of the weight matrices of the kernel modulator are trainable. We initialize the diagonal entries as: $\mathbf{U}^{(i,j)} = \mathbf{1}+\mathcal{N}(0, 0.001)$. This method implies that each input weight is modulated by a single multiplicative scaling parameter instead of combining neighboring input values. 

Compared to the default initialization, the orthogonal method leads to a similar classification accuracy on CIFAR-10, which means this method could be an alternative to the default method.
On first sight, the diagonal method also keeps the input values as in the default method. However, the accuracy is 15\% worse than the model initialized using the default initialization, suggesting that using the fully-connected kernel modulator is critical for achieving better accuracy.

\paragraph{Number of layers} We looked at six different network depths ranging from 1 to 50 layers in the kernel modulator (Fig.~\ref{fig:select:configs}C). When the number of layers increases past 2, the accuracy decreases while the standard deviation increases. Because the kernel modulator treats the channels of a convolution kernel independently, including more layers in the kernel modulator does not translate to higher accuracy. The results indicate that a shallow kernel modulator, \emph{e.g.,} one or two layers, is sufficient for reaching good accuracy.

\section{Discussion and Conclusion}\label{sec:conclusion}

This work introduces Kernel Modulation, a parameter-efficient method that trains a ConvNet through the combination of explicit and implicit modulations.
KM offers a simple strategy for training and deploying neural networks to meet the growing number of tasks. For example, distributing 100 task-specific ResNet-50 networks would require $100\,\text{tasks}\times94\,\text{MB/task-specific network}=9400\,\text{MB}$ for the network weights. Each network update via KM constitutes on average, 1.4\% of parameters for one task. Therefore, the memory requirement is reduced to $94\,\text{MB}+100\,\text{tasks}\times94\,\text{MB/task-specialized network}\times1.4\%=226\,\text{MB}$, a 43X memory reduction.

We evaluated the KM model in three sets of experiments using seven benchmark datasets. The results showed that the KM model performed better or competitive network accuracy when compared with previous parameter-efficient methods, \emph{e.g.}, 9\% higher accuracy in the Transfer Learning experiments. In addition, the analysis of different network configurations of the MLP kernel modulators on the CIFAR-10 classification task showed that the best kernel modulator is a 2-layer MLP network that uses $\tanh$ activation function. The initialization preserves the range of the input weight values using a superposition of an identity matrix and a normal distribution.

In this work, the kernel modulator treats weight channels as independent inputs, therefore it does not utilize neighboring weight channels to determine the modulated weights.
A future direction would be to replace the MLP network with other candidates that use neighboring weight channels.
Additionally, one could consider modulation of only 
convolution layers that require large weight changes during training, \emph{i.e.}, ``critical'' layers instead of applying the kernel modulator to all convolution layers as in this work.

Kernel Modulation facilitates and promotes cost-efficient on-device ConvNet deployment by reducing the memory footprint when distributing task-specialized networks.

\bibliographystyle{IEEEtran}
\bibliography{IEEEabrv, icpr22}
%



\end{document}